\documentclass[10pt,twocolumn, letterpaper]{article}
\usepackage{iccv}
\usepackage{times}
\usepackage{epsfig}
\usepackage[dvipsnames]{xcolor}
\usepackage{colortbl}
\usepackage{graphicx}
\usepackage{amsmath}
\usepackage{amssymb}
\usepackage{booktabs}
\usepackage{amssymb}
\usepackage{amsmath}
\usepackage{color}
\usepackage{multirow}
\usepackage{amssymb}

\usepackage{arydshln}

\usepackage{adjustbox}
\usepackage{bm}

\usepackage[pagebackref=true,breaklinks=true,letterpaper=true,colorlinks,citecolor=blue,linkcolor=blue,bookmarks=false]{hyperref}
\iccvfinalcopy 

\def\iccvPaperID{11039} 

\ificcvfinal\fi

\begin{document}

\title{Generative Multiplane Neural Radiance for 3D-Aware Image Generation}

\author{Amandeep Kumar\textsuperscript{1} \hspace{.1cm} Ankan Kumar Bhunia\textsuperscript{1} \hspace{.1cm} Sanath Narayan\textsuperscript{2} \hspace{.1cm} Hisham Cholakkal\textsuperscript{1} \\ \hspace{.1cm} Rao Muhammad Anwer\textsuperscript{1,3}  Salman Khan\textsuperscript{1} Ming-Hsuan Yang \textsuperscript{4, 5, 6} Fahad Shahbaz Khan\textsuperscript{1,7} \hspace{.1cm} 
\\
\textsuperscript{1}Mohamed bin Zayed University of AI \hspace{.1cm} \textsuperscript{2}Technology Innovation Institute
\hspace{.1cm}   \textsuperscript{3}Aalto University  \hspace{.1cm} \\ \textsuperscript{4}University of California, Merced \hspace{.1cm} 
\textsuperscript{5}Yonsei University \hspace{.1cm} 
\textsuperscript{6}Google Research \hspace{.1cm} 
\textsuperscript{7}Link{\"o}ping University\\
}

\maketitle
\ificcvfinal\thispagestyle{empty}\fi


\begin{abstract}
We present a method to efficiently generate 3D-aware high-resolution images that are view-consistent across multiple target views. The proposed multiplane neural radiance model, named GMNR, consists of a novel $\bm\alpha$-guided view-dependent representation ($\bm\alpha$-VdR) module for learning view-dependent information. 
%
%
The $\bm\alpha$-VdR module, faciliated by an $\bm{\alpha}$-guided pixel sampling technique, computes the view-dependent representation efficiently by learning  viewing direction and  position coefficients. 
Moreover, we propose a view-consistency loss to enforce photometric similarity across multiple views. 
The GMNR model can generate 3D-aware high-resolution images that are view-consistent across multiple camera poses, while
maintaining the computational efficiency in terms of both training and inference time. 
Experiments on three datasets demonstrate the effectiveness of the proposed modules, leading to favorable results in terms of both generation quality and inference time, compared to existing approaches. 
Our GMNR model generates 3D-aware images of $1024 \times 1024$ pixels with $17.6$ FPS on a single V100. 
Code : \url{https://github.com/VIROBO-15/GMNR}

\end{abstract}

\section{Introduction}
\label{sec:intro}

The advances in generative adversarial networks (GANs) \cite{Goodfellow2014GenerativeAN} have resulted in significant progress in the task of high-resolution photorealistic 2D image generation \cite{Karras2021AliasFreeGA, karras2019style, Karras2020AnalyzingAI, kumar2021udbnet}. 
The problem of generating 3D-aware images that render an object in different target views has received increasing interest in the recent years. 
Learning such 3D-aware image generation is challenging due to the absence of 3D geometry supervision or multi-view inputs during training. 
Furthermore, the synthesized 3D-aware images are desired to be of high-resolution, generated at extrapolated views (\ie, large non-frontal views) and consistent across camera views.

In the absence of 3D supervision, existing 3D-aware image generation approaches \cite{Chan2022EfficientG3, Deng2022GRAMGR} typically rely on learning the 3D geometric constraints by using either implicit \cite{Schwarz2020GRAFGR, Niemeyer2021GIRAFFERS, Gu2022StyleNeRFAS, OrEl2022StyleSDFH3} or explicit \cite{lombardi2019neural, Sitzmann2019DeepVoxelsLP} 3D-aware inductive biases and a rendering engine. 
While implicit representations, \eg, neural radiance fields \cite{Mildenhall2020NeRFRS} (NeRF),  possess the merits of better handling complex scenes along with memory efficiency, their slow querying and sampling generally negatively affects the training duration, inference time as well as the 3D-aware generation of high-resolution images.
On the other hand, explicit representations, \eg,  voxel grid \cite{Schwarz2022VoxGRAFF3},  are typically fast but have large memory footprint leading to scaling issues at higher resolutions.
These issues are recently addressed \cite{zhao2022generative} by utilizing multiplane images (MPI) as an explicit representation to transfer the knowledge learned by a 2D GAN to 3D-awareness. 
In this way, existing 2D GANs, \eg, StyleGAN \cite{karras2019style}, can be extended to obtain the alpha maps conditioned on the plane's depth, followed by conditioning the discriminator on a target pose for training the image synthesis model. 

While the aforementioned scheme of avoiding a volumetric rendering of pixels enables efficient training and inference, it may lead to inaccurate rendering of object shapes at extrapolated views due to fewer multiplanes during training. 
Moreover, inconsistent artifacts across different views can occur since such a scheme optimizes the warping from canonical pose to a single target pose.
A straightforward way to overcome this issue is to increase the resolution in the disparity space, \ie, more planes. 
This can likely help in reducing these artifacts resulting in improved rendering at extrapolated viewing angles. 
However, this will result in significantly increasing the training time  as well as memory overhead. 
In this work, we show how to collectively address the above issues without any significant degradation of training and inference speed.

\begin{figure*}[t!]
\centering
\includegraphics[width=0.97\textwidth]{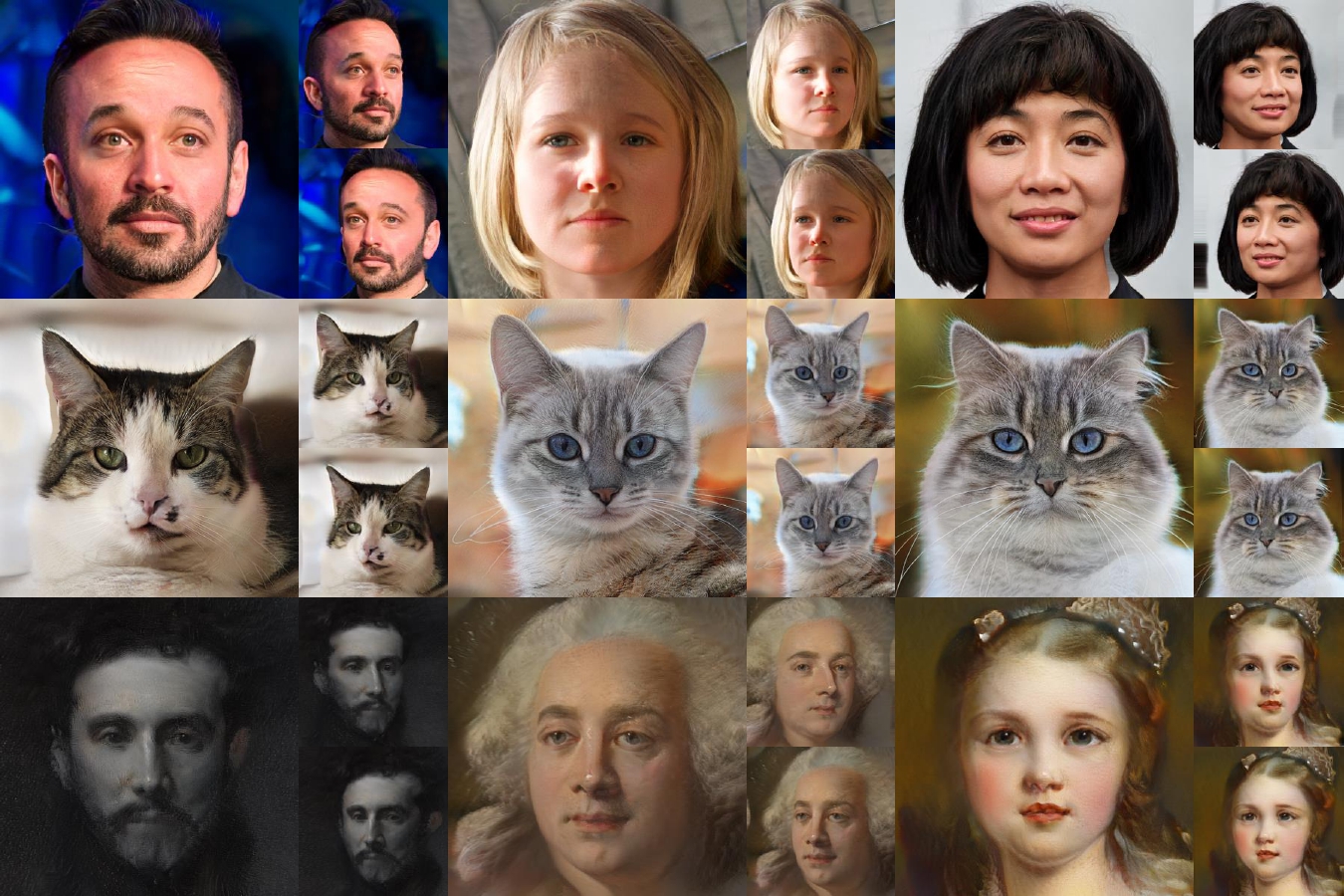}
\caption{Generated examples using our proposed 3D-aware view-consistent GMNR approach. For each example, we show the generated canonical view along with the rendered images at two different target poses. Our GMNR efficiently synthesizes 3D-aware high-resolution ($512 \times 512$ in row $2$; $1024 \times 1024$ in rows $1$ and $3$) scenes with detailed geometry along with consistent rendering across multiple views at a speed of $17.6$ frames per second ($1024 \times 1024$ pixels) on a single Tesla V100 GPU. \vspace{-0.3cm}
 }
\label{fig:intro}
\end{figure*}

\noindent\textbf{Contributions:} We propose an efficient approach named Generative Multiplane Neural Radiance (GMNR), that learns to synthesize 3D-aware and view-consistent high-resolution images across difference camera poses. 
To this end, we introduce a novel $\bm\alpha$-guided view-dependent representation module ($\bm\alpha$-VdR) that enables the generator to better learn view-dependent information during training. 
Our $\alpha$-VdR employs a linear combination of learnable \textit{image-specific} viewing direction and \textit{image-agnostic} position coefficients along with an $\bm{\alpha}$-guided pixel sampling technique to compute the view-dependent representation efficiently. 
The proposed sampling technique ensures that a balanced set of valid pixel locations from each multiplane is considered when computing the view-dependent representation, resulting in 3D-aware high-resolution images with diminished artifacts in the target poses. 
Moreover, we employ a view-consistency loss for enforcing photometric similarity across multiple rendered views. 
Consequently, our GMNR generates 3D-aware high-resolution images that are view consistent across different camera poses while maintaining the computational efficiency at inference.

Extensive qualitative and quantitative experiments are conducted on three datasets: FFHQ \cite{Karras2019ASG}, AFHQv2-Cats \cite{Choi2020StarGANVD} and MetFaces \cite{Karras2020TrainingGA}. 
Our GMNR performs favorably against existing works published in literature. 
When generating images of $1024 {\times} 1024$ pixels on FFHQ dataset, GMNR outperforms the best existing approach~\cite{zhao2022generative} by reducing the FID from $7.50$ to $6.58$, while operating at a comparable inference speed of $17.6$ frames per second (FPS) on a single tesla V100. 
Fig.~\ref{fig:intro} shows 3D-aware high-resolution generated scenes from our GMNR exhibiting detailed geometry and consistent rendering across multiple views.

\section{Preliminaries}

\noindent\textbf{Problem Statement:}
In this work, the goal is to learn a 2D GAN for generating 3D-aware high-resolution images that are view-consistent, such that the generated images identically encapsulate the synthesized objects at different target camera poses $p_t$. 
Here, multiplane images are generated to capture the 3D information and are then utilized to render a view-consistent 2D image at a target camera pose $p_t$. 
The multiplane images consist of a set of $L$ fronto-parallel planes $i \in \{1, \cdots, L\}$, each of size $H \times H \times 4$, \ie, each plane $i$ comprises an RGB image $C_i \in \mathbb{R}^{H \times H \times 3}$ and an alpha map $\bm\alpha_{i} \in [0, 1]^{H \times H \times 1}$. 
The distance between the camera and a plane $i$ is denoted by depth $d_i\in \mathcal{R}$. 
Next, we describe our baseline framework for generating view-consistent 3D-aware high-resolution images.

\subsection{Baseline Framework}
Our baseline model is motivated by the recent multiplane image generation method, GMPI~\cite{zhao2022generative}, since it focuses on computationally efficient generation of 3D-aware images.
%
GMPI extends the StyleGANv2~\cite{Karras2020AnalyzingAI} network with a branch for obtaining alpha maps and a differentiable renderer for generating 3D images at different target camera poses. 
Moreover, the base GMPI framework reuses the same color-texture across all planes, in turn reducing the task of StyleGANv2 generator $f_G(\cdot)$ to synthesizing a single RGB image $C$ and the corresponding per-plane alpha maps $\bm\alpha_i$, given by
%
%
\begin{equation}
\bm{M} \triangleq \{C, \{\bm\alpha_1, \cdots, \bm\alpha_L\}\} = f_G(z, \{d_1, \cdots, d_L\}),
\end{equation}
where $z$ denotes the latent vector input to $f_G(\cdot)$. 
For generating the alpha maps at a resolution $r \in \{4, 8, \cdots \}$, a single convolutional layer $f_{ToAlpha}^r$ is first used to generate the $\hat{\bm\alpha}_i^r$ from intermediate feature representations $F_{\bm\alpha_i}^r$, given by
\begin{align}
    & \hat{\bm\alpha}_{i}^r
        = f_{ToAlpha}^r (F_{\bm\alpha_i}^r), \\
    & F_{\bm\alpha_i}^r = \frac{F^r - \mu(F^r)}{\sigma(F^r)} + f_{Emb}(d_i, e),
\end{align}
where $\sigma(F^r)$, $\mu(F^r)\!\in\! \mathbb{R}^{dim_r }$ denote the standard deviation and mean of the feature $F^r \!\in\! \mathbb{R}^{r \times r \times dim_r}$. 
The plane specific embedding $f_{Emb}(d_i,e)$ is computed using the style embedding $e$ and depth $d_i$ of plane $i$, similar to StyleGANv2. 
Note that $F_{\bm\alpha_i}^r$ is specific to each plane, while $f_{ToAlpha}^r$ is shared across all planes. 
Finally, the alpha maps $\bm\alpha_i^r$ at different resolutions $r$ are accumulated through an upsampling operation that is consistent with the StyleGANv2 design. 
With this formulation, GMPI introduces a branch to generate alpha maps conditioned on the plane depths $d_i$ by utilizing the intermediate feature representations and conditions the discriminator on the camera poses to make the 2D StyleGANv2 3D-aware~\cite{zhao2022generative}.
%

\begin{figure}[t!]
  \begin{center}
    \includegraphics[width=0.98\columnwidth]{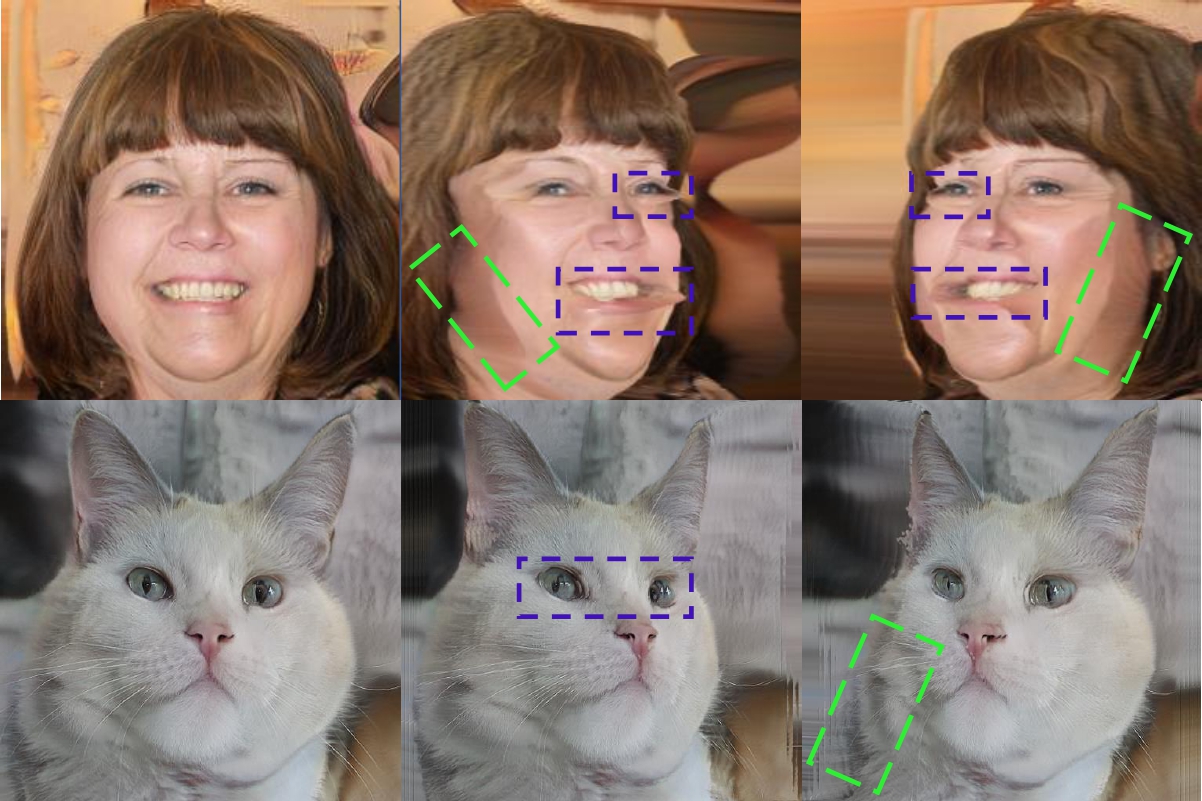}
    \caption{Example generated images using the baseline approach depicting the frontal (canonical) and target views. Here, the baseline model is trained on FFHQ (rows 1) and AFHQv2-Cats (row 2). While being effective in synthesizing frontal views (col. 1), the baseline struggles when generating the target views and introduces artifacts  during the rendering (col. 2 and 3). In these cases, the generated images depict repeated textures (highlighted as blue box) due to the same RGB content in each plane and layered artifacts (highlighted as green box) due to fewer planes employed during training. For more examples, see supplementary material. \vspace{-0.5cm}
    }
    \label{fig:failure}
  \end{center}
\end{figure}

Our baseline framework extends the existing 2D StyleGANv2 to make it 3D-aware using  implict and explicit representations.
Moreover, by first generating multiplane images at canonical view and then warping them to target poses, it avoids a volumetric rendering leading to an efficient training and inference. 
However, the baseline model does not  effectively render object images at extrapolated views, likely due to fewer multiplanes employed to overcome memory issues during training. 
Furthermore, since the baseline optimizes by warping from canonical view to a single random target pose, it leads to inconsistent artifacts across multiple views (see Fig.~\ref{fig:failure}). 
Next, we present our approach that collectively address the above issues without any significant change in training and inference time.

\begin{figure*}[t!]
\centering
\includegraphics[width=1\textwidth]{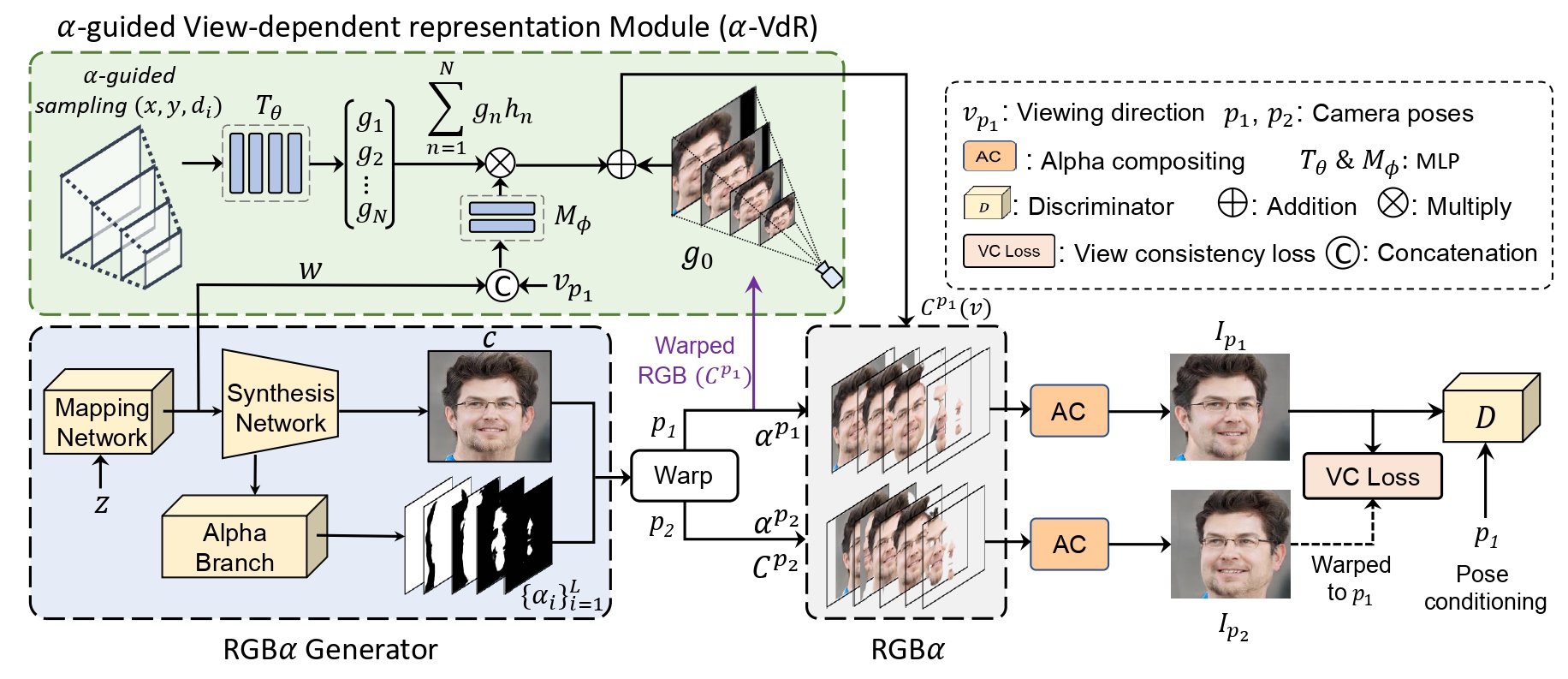}\vspace{-0.25cm}
\caption{Overall architecture of our GMNR for generating 3D-aware and view-consistent images at high-resolution. Our GMNR takes a latent vector $z \in \mathbb{R}^{d_z}$ and outputs an RGB image at a target view. GMNR comprises an RGB$\alpha$ generator, an $\bm\alpha$-guided view-dependent representation ($\bm\alpha$-VdR) module, a differentiable renderer and a pose-conditioned discriminator. The RGB$\alpha$ generator synthesizes the RGB image and the alpha maps $\{\bm\alpha_i\}_{i=1}^L$ corresponding to the canonical pose. The generated RGB image that is warped to a target pose $p_1$ is then input along with the style-code $\bm{w}$ to the $\bm\alpha$-VdR module (Sec.~\ref{sec:VdR}). 
The $\bm\alpha$-VdR module learns a view-dependent pixel representation using a linear combination of coefficients (image-agnostic $\{g_n\}_{n=1}^N$ and image-specific $\{h_n(\bm{w})\}_{n=1}^N$) computed from the $\bm\alpha$-guided sampling positions $(x,y,d_i)$, viewing direction $v_{p_1}$ and style-code $\bm{w}$ using two MLP networks $T_\theta(\cdot)$ and $M_\phi(\cdot)$, respectively. Here, the $\alpha$-guided pixel sampling aids in efficiently sampling the pixel positions for computing the view-dependent representation. As a result, $\bm\alpha$-VdR module learns to model the image-specific view-dependent 3D characteristics.
Moreover, a view-consistency loss $\mathcal{L}_{vc}$ (Sec.~\ref{sec:vc_loss}) is employed for enhancing the photometric consistency across different views of the rendered images. Consequently, the RGB$\alpha$ generator along with the $\bm{\alpha}$-VdR module and renderer synthesize 3D-aware view-consistent images at target poses during inference.\vspace{-0.2cm}}
\label{fig:main}
\end{figure*}

\section{Proposed Approach\label{sec:approach}}
\noindent\textbf{Overall Architecture:}
Fig.~\ref{fig:main} presents the overview of our proposed GMNR framework. 
Within our GMNR, the RGB$\bm\alpha$ generator adapts a conventional 2D generator by integrating an $\bm\alpha$-branch with StyleGANv2, yielding a set of fronto-parallel alpha maps $\{\bm\alpha_i\}_{i=1}^L$, as in the baseline. 
The focus of our design is the introduction of an \textit{$\bm\alpha$-guided view-dependent representation} ($\bm\alpha$-VdR) that learns image-specific view-dependent information, crucial for rendering images with diminished artifacts in target poses.
The $\bm\alpha$-VdR module learns a view-dependent pixel representation using a linear combination of coefficients obtained from two MLPs by efficiently sampling pixel positions through an $\bm\alpha$-guided pixel sampling technique. 
This enables the $\bm\alpha$-VdR module to learn modeling the image-specific view-dependent 3D characteristics.
Moreover, we employ a \textit{view-consistency loss} to enforce photometric consistency across different views of the rendered images. 
Consequently, 3D-aware view-consistent images of high-resolution at target poses are synthesized by the  RGB$\alpha$ generator together with the $\bm\alpha$-VdR module and renderer at inference. 

\subsection{$\bm\alpha$-guided View-dependent Representation}
\label{sec:VdR}


As discussed earlier, the baseline model renders 3D images using an MPI representation without explicitly utilizing the view-dependent information during training, which is desired for 3D-aware view-consistent image generation. 
To learn view-dependent information, we introduce an $\bm\alpha$-guided view-dependent representation module ($\bm\alpha$-VdR) that comprises two separate MLP networks $T_\theta(\cdot)$ and $M_\phi(\cdot)$. 
We consider a pixel to be a discrete sample of a radiance function $R(q,v)$ with $q, v \in \mathbb{R}^3$ as the pixel coordinate and the target viewing direction. 
Motivated by \cite{kautz1999hardware,kautz1999interactive}, we note that $R(q,v)$ can be approximated by a sum of products $g_{n}\cdot h_n$. Here, $g_n$ and $h_n$ are computed using two MLP networks $T_\theta(\cdot)$ and $M_\phi(\cdot)$ with inputs $q$ and $v$, respectively.

Given the pixel location $q=(x,y,d_i)$ for a plane depth $d_i$ ($i\in\{1,\cdots,L\}$), the MLP $T_\theta(\cdot)$ predicts the \textit{image-agnostic} position coefficients $\{g_1^q,\cdots,g_N^q\}$. Similarly, to enable \textit{image-specific} view-dependent modeling, the normalized viewing direction $v = (v_x, v_y, v_z)$ is utilized along with the style-code $\bm{w}$ generated by the mapping network of the StyleGANv2 in the RGB$\alpha$ generator. 
To this end, $v$ and $\bm{w}$ are concatenated and  input to the MLP $M_\phi(\cdot)$, which outputs image-specific viewing direction coefficients $\{h_1^v(\bm{w}),\cdots,h_N^v(\bm{w})\}$. 
The color representation $s^q(v)$ of a pixel $q$ for a target viewing direction $v$ is computed using
\begin{equation}
\label{eq:color_repr}
s^q(v) = g_{0}^q + \sum_{n=1}^N g_{n}^{q}\cdot h_n^v(\bm{w}).
\end{equation}
Note that $g_0$ is computed by performing a homography warping operation on the RGB image generated at canonical pose to the target camera pose $p_1$.

\begin{figure}[t!]
  \begin{center}
    \includegraphics[width=0.98\columnwidth]{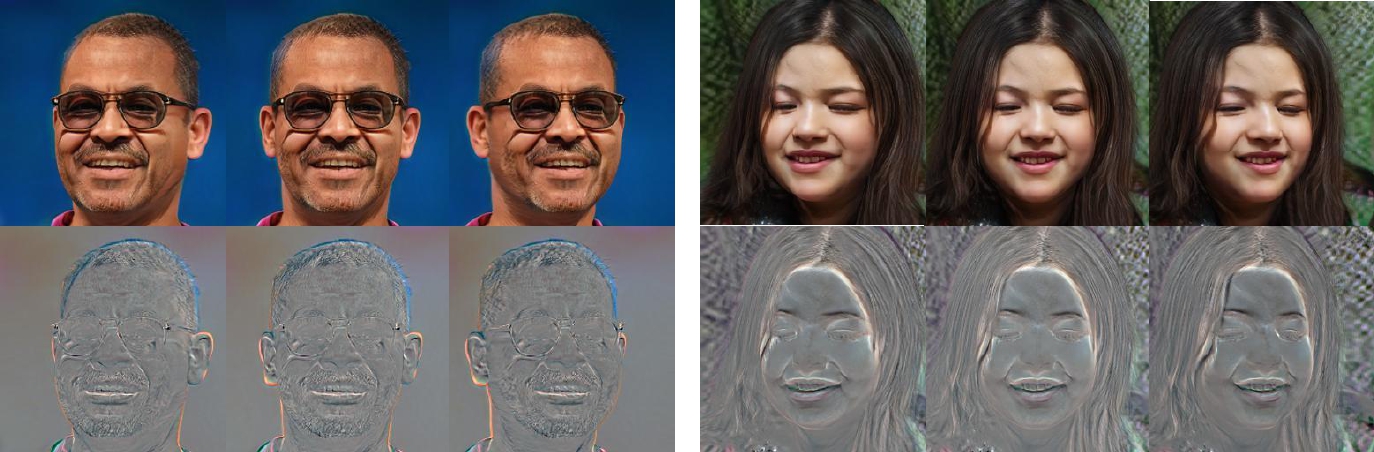}
    \caption{{Visualization of view-dependent information integrated by $\bm\alpha$-VdR module (bottom) at various poses of two example generated images (top)}. \vspace{-0.5cm}}
    \label{fig:vdr_rep}
  \end{center}
\end{figure}

\begin{figure}[t!]

  \centering
   
    \includegraphics[trim={0 0 0 0.9cm},clip, width=1\columnwidth]{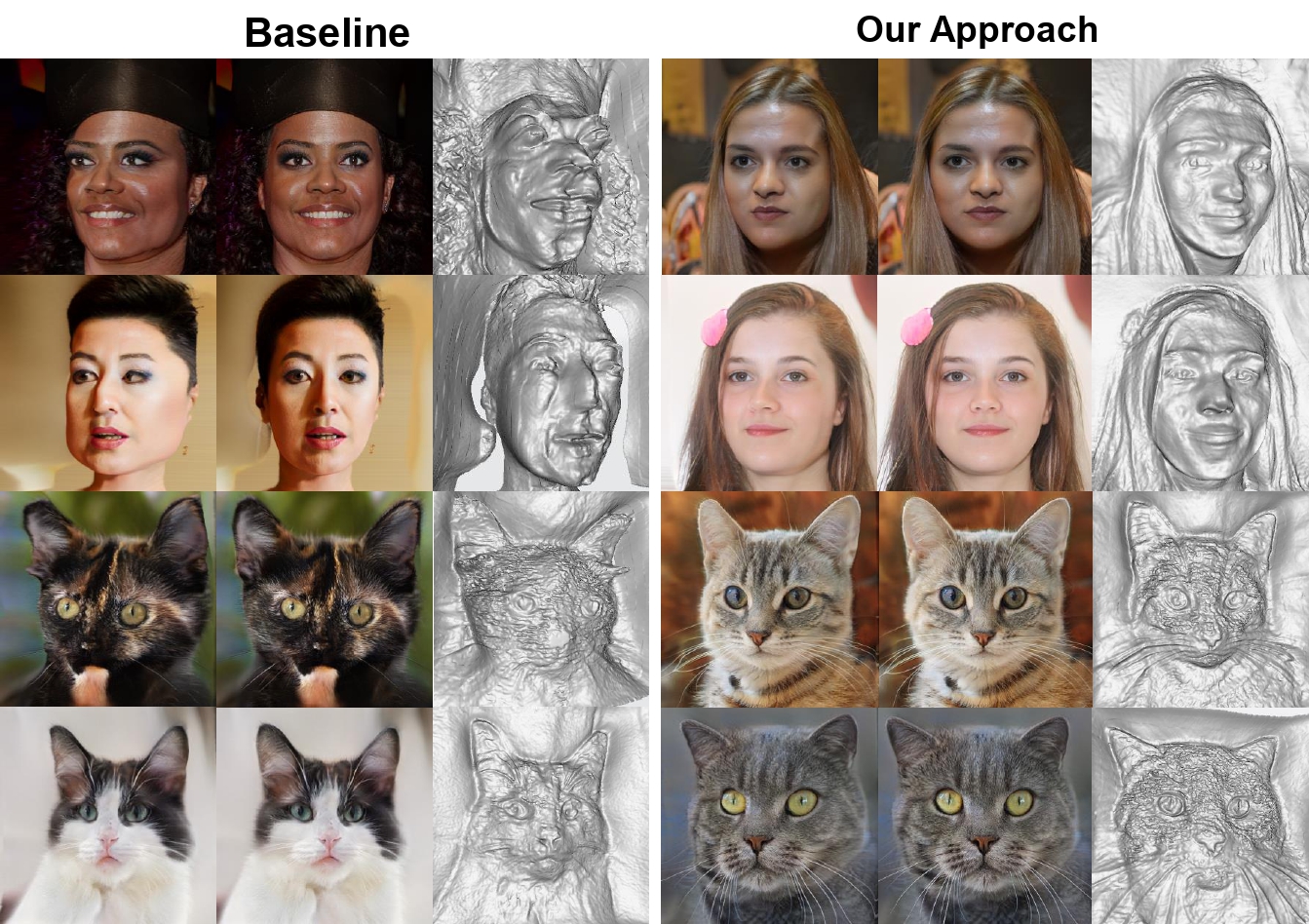}
$\hbox{\small \textbf{Baseline} \qquad  \qquad \quad \qquad      \vspace{0.2cm}
   \small \textbf{Our Approach}}$
    \caption{Synthesized rendered RGB image at a target view (left), its corresponding canonical view (center) along with the mesh (right) for the baseline and our GMNR, respectively. Compared to the baseline, our GMNR better renders the target objects at large non-frontal views due to learning the view-dependent information through the $\bm\alpha$-VdR module. Stretched eyes can be observed in the case of  baseline generated image (row 1, col. 1), while such an artifact is mitigated by our approach (row 1, col. 4).
    \vspace{-0.3cm} 
    }

    \label{fig:failure-vdr}
\end{figure}

\noindent\textbf{$\bm{\alpha}$-guided Pixel Sampling:}
\label{sec:sampling}
Sampling all the pixels for color representation learning (Eq.~\ref{eq:color_repr}) to generate \textit{high-resolution} images is cost-prohibitive, since it is infeasible to compute the view-dependent representation $s^q(v)$ of all the pixels $q$ in an MPI. 
To alleviate this issue, we introduce a new sampling technique with the aid of the generated alpha maps $\{\bm\alpha_{i}\}_{i=1}^L$, which reduces the volume of points that are required for sampling while retaining the fine details for coefficient learning. To this end, we compute a weight matrix $\bm{A}_i \in \mathbb{R}^{H\times H}$ corresponding to the plane $i$, given by
\begin{equation}
\label{eq:alpha_weight}
\bm{A}_i = \bm{\alpha}_i^{p_1}\cdot \prod_{j=1}^{i-1}(1-\bm\alpha_{j}^{p_1}),
\end{equation}
where $\bm\alpha_i^{p_1}$ denotes $\bm\alpha_i$ warped to target pose $p_1$. A pixel location $(x,y,d_i)$ is a candidate for sampling if the corresponding weight is greater than $0$, \ie, if $\bm{A}_i(x,y) > 0$. Note that Eq.~\ref{eq:alpha_weight} ensures that a pixel location $(x,y,d_i)$ is not considered during sampling if either $\bm\alpha_i^{p_1}(x,y) = 0$ or if $\bm\alpha_j^{p_1}(x,y) = 1$ for any $j<i$. Furthermore, among the candidate locations $(x,y,d_i)$, a \textit{per-plane} random sampling is performed to select only a certain percentage of the candidate locations in a plane $i$. 
This balanced sampling strategy guided by alpha maps ensures that every plane $i$ is adequately represented during the color representation computation (Eq.~\ref{eq:color_repr}), thereby leading to effective learning of view-dependent information. 
As such, our $\bm\alpha$-VdR module outputs a view-dependent RGB image $C^{p_1}(v)$ at a target pose $p_1$ and incorporates 3D-awareness during image generation. Fig.~\ref{fig:vdr_rep} shows the view-dependent information integrated by $\bm\alpha$-VdR module when rendering objects at different views. 
Moreover, Fig.~\ref{fig:failure-vdr} shows the impact of $\bm\alpha$-VdR module on generated examples from FFHQ and AFHQv2-Cats. Compared to the baseline, our approach with $\bm\alpha$-VdR module  synthesizes high-resolution images at target views with diminished artifacts.

\subsection{View-consistency Loss}
\label{sec:vc_loss}
To achieve a better photometric consistency across multiple views, we employ an image-level optimization loss. Let $I_{p_1}$ and $I_{p_2}$ denote the images rendered at target poses $p_1$ and $p_2$. Note that only $I_{p_1}$ is generated by integrating the view-dependent information from $\bm\alpha$-VdR module, while $I_{p_2}$ is rendered directly from the RGB$\alpha$ generator output, as shown in Fig.~\ref{fig:main}. We first warp $I_{p_2}$ to target pose $p_1$ to obtain the warped image $I_{\psi}$. Afterwards, we employ the image-level optimization loss that enhances the view-consistency between the primary image $I_{p_1}$ and the warped image $I_{\psi}$ in order to satisfy the geometry requirements between views. Similar to the image reconstruction problem \cite{godard2019digging, zhou2017unsupervised}, we formulate the image-level optimization loss as a combination of SSIM \cite{wang2004image} and $\ell_1$ \cite{zhao2016loss}, given by
\begin{equation}
 \mathcal{L}_{vc} = \frac{\delta}{2}(1 - SSIM(I_{p_1}, I_{\psi})) + (1 - \delta)||I_{p_1} - I_{\psi}||_1 .
\end{equation}
%
%

\subsection{Training and Inference}
\noindent\textbf{Training:}
Our GMNR framework is trained using a pose conditioned discriminator $D$, which compares the fake images generated by the RGB$\alpha$ generator and real training images $I_{gt}$. The overall loss formulation, utilizing a non-saturating GAN loss with $R1$ penalties \cite{mescheder2018training} and the view-consistency loss $\mathcal{L}_{vc}$ is given by
\begin{align}
 \mathcal{L} =& \mathbb{E}_{I_{p_{t}}, p_t}[f(\log Q(I_{p_t}, p_t))]  + \mathbb{E}_{I_{gt}, p_t} [f(\log Q(I_{gt}, p_t ))] \nonumber \\
 + &  \eta|\nabla_{I_{gt}} \log P(y=\text{real}|I_{gt}, p_t )|^2] + \lambda  \mathcal{L}_{vc},
\end{align}
where $Q(I, p_t) =  P(y=\text{real} |I, p_t )$ denotes the probability that image $I$ from a camera pose $p_t$ is real, $f(x) = -\log(1+\exp(-x))$ and $\eta = 10.0$.\\
%
%
\noindent\textbf{Inference:}
During inference, we use the RGB$\alpha$ generator to synthesize RGB image $C$ and alpha maps $\bm\alpha_i$. The $\bm\alpha$-VdR module takes the semantic information $\bm{w}$ generated by the StyleGANv2 along with target viewing direction $v_t$ as inputs, and computes an image-specific view-dependent representation $C^{p_t}(v_t) \in \mathbb{R}^{H\times W \times 3 \times L}$ (based on Eq.~\ref{eq:color_repr}). This $C^{p_t}(v_t)$ together with $\bm\alpha_i$ are used to obtain multiplane images, which are then input to the MPI renderer. The renderer warps them to the target pose $p_t$ followed by alpha composition for combining the planes to obtain the desired 3D-aware and view-consistent image $I_{p_t}$. Note that while the image-specific coefficients $\{h_n(\bm{w})\}$ in Eq.~\ref{eq:color_repr} are computed once for an image, the image-agnostic coefficients $\{g^q_n\}$ are computed only once, leading to minimal computational overhead during inference.


\begin{table*}[t!]
\centering
\caption{Comparisons between the baseline and our GMNR on FFHQ and AFHQv2-Cats. We also report the training time when utilizing $8$ Tesla V100 GPUs. Our GMNR achieves consistent improvement in performance on all metrics and different resolutions, compared to the baseline. Furthermore, this improvement in performance over the baseline is achieved without any significant degradation in the training time and the inference speed of the model.  }
\label{tab:comp_base}
\setlength{\tabcolsep}{2.5pt}
\adjustbox{width=0.75\textwidth}{
\begin{tabular}{llcccccccccccc} 
\toprule[0.1em]
\multirow{2}{*}{ } & \multirow{2}{*}{\textbf{Method}}   & \multicolumn{6}{c}{\textbf{FFHQ}}                    & & \multicolumn{3}{c}{\textbf{AFHQv2-Cats}}                                                                                                                   \\ 
\cmidrule{3-8} \cmidrule{11-12}
                      &                           
                      & $FID$$\downarrow$ & $KID$ $\downarrow$ & $ID$$\uparrow$  & $Depth$$\downarrow$  & $Pose$$\downarrow$  & Train Time$\downarrow$ & Infer Speed$\uparrow$                          
                      & & $FID$$\downarrow$ & $KID$ $\downarrow$                             
                                                           \\ 
\toprule[0.1em]

\multirow{2}{*}{$256^2$}  & {Baseline}                      & $11.4$     & $0.738$          & $0.700$ & $0.53$ & $0.0040$  & $3$h &  $328$ FPS &  & n/a   & n/a  \\
& \cellcolor{cyan!25} {\textbf{Ours: GMNR}} & \cellcolor{cyan!25}$ \cellcolor{cyan!25} 9.20$      & \cellcolor{cyan!25}$ \cellcolor{cyan!25} 0.720$  & \cellcolor{cyan!25} $0.730$ & \cellcolor{cyan!25}$0.39$ & \cellcolor{cyan!25}$0.0032$ & \cellcolor{cyan!25}$3$h $33$m &  \cellcolor{cyan!25}$313$ FPS & \cellcolor{cyan!25}& \cellcolor{cyan!25}n/a   & \cellcolor{cyan!25}n/a  \\

\hdashline
\multirow{2}{*}{$512^2$}   & {Baseline}                      & $8.29$     & $0.454$          & $0.740$ & $0.46$ & $0.0060$  & $5$h & $83.5$ FPS & & $7.79$   & $0.474$  \\
                           & \cellcolor{cyan!25} {\textbf{Ours: GMNR}}& \cellcolor{cyan!25}$6.81$  & \cellcolor{cyan!25}$0.370$   & \cellcolor{cyan!25}$0.760$ & \cellcolor{cyan!25}$0.40$ & \cellcolor{cyan!25}$0.0052$   & \cellcolor{cyan!25}$5$h $42$m  & \cellcolor{cyan!25}$78.9$ FPS & \cellcolor{cyan!25}& \cellcolor{cyan!25}$6.01$   & \cellcolor{cyan!25}0.450  \\                            

\hdashline
\multirow{2}{*}{$1024^2$}  & {Baseline}                      & $7.50$     & $0.407$          & $0.750$ & $0.54$ & $0.0070$  & $11$h  & $19.4$ FPS &  & n/a   & n/a  \\
                           & \cellcolor{cyan!25}{\textbf{Ours: GMNR}} & \cellcolor{cyan!25}$6.58$ & \cellcolor{cyan!25}$0.351$  & \cellcolor{cyan!25}$0.769$   & \cellcolor{cyan!25}$0.43$  & \cellcolor{cyan!25}$0.0064$  & \cellcolor{cyan!25}$12$h  & \cellcolor{cyan!25}$17.6$ FPS & \cellcolor{cyan!25}& \cellcolor{cyan!25}n/a   & \cellcolor{cyan!25}n/a  \\ 
\bottomrule[0.1em]
\end{tabular}
}
\end{table*}

\vspace{-0.1cm}
\setlength{\tabcolsep}{2.5pt}
\begin{table}[t]
\centering
\caption{Effect of progressively integrating our proposed contributions into the baseline on FFHQ dataset with {$512^2$} resolution. The introduction of the proposed $\bm\alpha$-VdR (Sec. \ref{sec:VdR}) into the baseline results in a consistent improvement in performance on all metrics. The results are further improved when integrating the proposed  view-consistency loss $\mathcal{L}_{vc}$ (Sec. \ref{sec:vc_loss}). }
\label{tab:comp_abla}

\resizebox{0.49\textwidth}{!}{
\begin{tabular}{llccccccccccc} 
\toprule[0.1em]

                      & \textbf{Method}
                      & $FID$$\downarrow$ & $KID$ $\downarrow$ & $ID$$\uparrow$  & $Depth$$\downarrow$  & $Pose$$\downarrow$                              
                                                 
                                                           \\ 
\toprule[0.1em]

                    & {Baseline }                     & $8.29$     & $0.454$          & $0.740$ & $0.457$ & $0.0060$ &    \\
                    
                    & Baseline + $\bm\alpha$-VdR     & $7.01$     & $0.381$          & $0.751$ & $0.412$ & $0.0054$         \\ 
                    & \cellcolor{cyan!25}{Baseline + $\bm\alpha$-VdR + $\mathcal{L}_{vc}$ }  & \cellcolor{cyan!25}$6.81$     & \cellcolor{cyan!25}$0.370$  & \cellcolor{cyan!25}$0.760$ & \cellcolor{cyan!25}$0.400$ & \cellcolor{cyan!25}$0.0052$    \\

\bottomrule[0.1em]
\end{tabular}
}
\end{table}

\section{Experiments}
\noindent\textbf{Datasets:} The proposed GMNR is evaluated on three datasets: FFHQ, AFHQv2 and MetFaces. The \textbf{FFHQ}~\cite{Karras2019ASG} dataset comprises $70,000$ high-quality images of real people's faces at $1024\times 1024$ resolution captured from various angles.
The \textbf{AFHQv2-Cats}~\cite{Choi2020StarGANVD, Karras2021AliasFreeGA} dataset has $5,065$ images ($512\times 512$ size) of cat faces at various angles. The \textbf{MetFaces}~\cite{Karras2020TrainingGA} dataset consists of $1,336$ high-quality face images extracted from the Metropolitan Museum of Art's collection. While an off-the-shelf pose estimator~\cite{deng2019accurate} is employed to compute a face's pose required for the pose conditioning for FFHQ and MetFaces, a cat face landmark predictor \cite{cathipsterizer} along with OpenCV \cite{opencv} perspective-n-point technique are used to compute the pose for AFHQv2-Cats images. Furthermore, horizontal flips are used as augmentation for AFHQv2-Cats and MetFaces. \\
\noindent\textbf{Evaluation Metrics:} In this work, five metrics are employed for quantitatively comparing the generated image quality, as in~\cite{zhao2022generative}. The Frechet Inception Distance (FID) \cite{heusel2017gans} and Kernel Inception Distance (KID) \cite{binkowski2018demystifying} are computed between 50$K$ generated images rendered at different random poses and (a) 50$K$ real images for FFHQ; (b) $5,065$ real images with flip augmentation for AFHQv2-Cats. The multi-view facial identity consistency (ID) is computed by first generating $1,024$ MPI representations and then employing the mean Arcface~\cite{deng2019arcface} cosine similarity score between pairs of rendered views at random poses for the same face. The depth accuracy (Depth) is measured as the MSE between the rendered depth and the pseudo ground-truth depth obtained from a pre-trained face reconstruction model~\cite{deng2019accurate}  on the face mask area. Similarly, the 3D pose accuracy (Pose) is computed by comparing the pose input used for rendering and the yaw, pitch and roll predicted by~\cite{deng2019accurate} for the rendered image.

\subsection{Implementation Details}
Within our GMNR, the MLP $T_\theta(\cdot)$ comprises $4$ fully-connected (FC) layers with hidden size of $384$, while $M_\phi(\cdot)$ has $3$ FC layers with hidden size $64$. Here, Leaky-ReLU activation is used in both MLPs. We add sinusoidal positional encodings to the pixel location and the viewing direction inputs, as in \cite{Mildenhall2020NeRFRS}. While the batch size is set to $32$ for AFHQv2-Cats, it equals $64$, $32$ and $16$ for FFHQ256, FFHQ512 and FFHQ1024, respectively.  We set $\delta$ in $L_{vc}$ loss to $0.85$ and $\lambda$ to $0.5$. For $\bm\alpha$-guided pixel sampling, $6\%$ of valid pixels are sampled from each plane for FFHQ ($256^2$, $512^2$ sizes) and AFHQv2-Cats, while it is $4\%$ for Metfaces and FFHQ ($1024 \times 1024$). The learning rate for training our GMNR is set to $ 2 \times 10^{-3}$. As in~\cite{zhao2022generative}, we use 32 planes during training and 96 for inference in all experiments. Near and far depth of the MPI are set as $0.95/1.12$ (FFHQ and Metfaces), $2.55/2.8$ (AFHQv2-Cats). Depth normalization is performed as in \cite{zhao2022generative}. Our model is trained using $8$ Tesla V100 GPUs using PyTorch-1.9 \cite{pytorch}. Our code and models will be publicly released. Further details are provided in the supplementary. 

\subsection{Experimental Results}
\noindent\textbf{Baseline Comparison:} We first present a quantitative and qualitative comparison of our GMNR approach with the baseline GMPI on both FFHQ and AFHQv2-Cats datasets. Tab.~\ref{tab:comp_base}
shows the comparison in terms of FID, KID, ID, Depth, Pose metrics and training time. As in the baseline GMPI~\cite{zhao2022generative}, the comparison is presented at three different resolutions: $256\times256$, $512 \times 512$ and $1024 \times 1024$. 
Compared to the baseline, our GMNR achieves consistent improvement in performance on all metrics, without any significant degradation in the training time and inference speed. 
In the case of $512^2$ resolution, the baseline obtains FID scores of $8.29$ and $7.79$ on FFHQ and AFHQv2-Cats datasets, respectively. In comparison, our GMNR achieves favorable performance with FID scores of $6.81$ and $6.01$, respectively. Similarly, GMNR obtains improved performance by reducing the Depth error from $0.46$ to $0.40$ and KID from $0.454$ to $0.370$ on the FFHQ dataset, compared to the baseline.  


\begin{table*}[t!]
\centering
\caption{Comparisons with existing approaches. Here, both GMPI and our approach use 96 planes during inference and without applying any truncation tricks \cite{brock2018large, karras2019style}. Further, the KID score is reported in KID$\times$100. In case the corresponding work does not report the results, we denote it as `-'.  We report the results of GIRAFFE, pi-GAN and LiftedGAN from the EG3D paper. Results reported on the entire AFHQv2 dataset instead of cats only are denoted by `*'. For $256 {\times} 256$ and $1024 {\times} 1024$ resolutions, no results are reported on AFHQv2-Cats for both GMPI and our GMNR since the corresponding pre-trained  StyleGANv2 checkpoints are unavailable. Compared to the recent GMPI, our GMNR achieves consistent improvement in performance, while running at comparable  inference speed (FPS). Particularly, GMNR better generates high-resolution ($1024 \times 1024$) images where most existing works struggle to operate demonstrating its flexibility.  Further, GMNR achieves consistent improvement on all metrics and significantly reduces the FID from $7.50$ to $6.58$ on FFHQ, compared to GMPI.}
\label{tab:compare_sota}
\vspace{-0.01cm}
\setlength{\tabcolsep}{4pt}
\adjustbox{width=0.77\textwidth}{
\begin{tabular}{llcccccccccc} 
\toprule[0.1em]
\multirow{2}{*}{ } & \multirow{2}{*}{\textbf{Method}} & \multirow{2}{*}{\textbf{Infer Speed}$\uparrow$}  & \multicolumn{5}{c}{\textbf{FFHQ}}                    & & \multicolumn{2}{c}{\textbf{AFHQv2-Cats}}                                                                                                                   \\ 
\cmidrule{4-8} \cmidrule{10-11}
                      &  &                         
                      & $FID$$\downarrow$ & $KID$ $\downarrow$ & $ID$$\uparrow$  & $Depth$$\downarrow$  & $Pose$$\downarrow$  &                        
                      & $FID$$\downarrow$ & $KID$ $\downarrow$                             
                                                           \\ 
\toprule[0.1em]

\multirow{10}{*}{$256^2$}   & {GIRAFFE \cite{Niemeyer2021GIRAFFERS}}                    & $250$  & $31.5 $        & $1.992$          & $0.64$ & $0.94$ & $0.0890$  & & $16.1$   & $2.723$  \\
                           & {pi-GAN $128^2$ \cite{Chan2021piGANPI}}                     & $1.63$ & $29.9 $        & $3.573$          & $0.67$ & $0.44$ & $0.0210$ &  & $16.0$   & $1.492$  \\
                           & {LiftedGAN \cite{Shi2021Lifting2S}}                     &  $25 $& $29.8 $        & -          & $0.58 $& $0.40$ & $0.0230$  & & -   & -  \\
                           & {GRAM \cite{Deng2022GRAMGR}}                      &$180$ & $29.8$         & $1.160$          & - & - & -     & & -   & -  \\
                           & {StyleSDF \cite{OrEl2022StyleSDFH3}}                   & -   & $11.5$         & $0.265$          & - & - & -    &  & $12.8^*$   & $0.447^*$  \\
                           & {StyleNeRF \cite{Gu2022StyleNeRFAS}}                    & $16$  & $8.00$         & $0.370$          & - & - & -    &  & $14.0^*$   & $0.350^*$  \\
                           & {CIPS-3D \cite{Zhou2021CIPS3DA3}}                     & - &$ 6.97$        & $0.287$          & - & - & -  && -   & -  \\
                           & {EG3D \cite{Chan2022EfficientG3}}                     &$36$ & $4.80$       & $0.149$          & $0.76$ & $0.31$ & $0.0050$      & & $3.88$   & $0.091$  \\
                           & {GMPI \cite{zhao2022generative}}                    & $328$  & $11.4$     & $0.738$          & $0.70$ & $0.53$ & $0.0040$    & & n/a   & n/a  \\
                           & \cellcolor{cyan!25} {\textbf{Ours: GMNR}}  & \cellcolor{cyan!25}$313$ & \cellcolor{cyan!25}$ 9.20 $     & \cellcolor{cyan!25}$0.720 $         &\cellcolor{cyan!25}$ 0.73$ & \cellcolor{cyan!25}$0.39$ &\cellcolor{cyan!25}$ 0.0032$    & \cellcolor{cyan!25}& \cellcolor{cyan!25}n/a   & \cellcolor{cyan!25}n/a  \\ 
\hdashline

\multirow{4}{*}{$512^2$}   & {EG3D \cite{Chan2022EfficientG3}}                     &$35$ & $4.70 $      & $0.132$          & $0.77$ & $0.39$ & $0.0050$     & & $2.77$   & $0.041$  \\
                            & {StyleNeRF \cite{Gu2022StyleNeRFAS}}                   &$14$   & $7.80$         & $0.220$          & - & - & -      & & $13.2^*$   & $0.360^*$  \\
                           & {GMPI \cite{zhao2022generative}}                    &$83.5$  & $8.29  $   & $0.454$          & $0.74$ &$ 0.46 $& $0.0060$    &  & $7.79$   & $0.474$  \\
                           & \cellcolor{cyan!25} {\textbf{Ours: GMNR}} &\cellcolor{cyan!25}$78.9$ & \cellcolor{cyan!25}$6.81$     & \cellcolor{cyan!25}$0.370 $         & \cellcolor{cyan!25}$0.76$ & \cellcolor{cyan!25}$0.40$ & \cellcolor{cyan!25}$0.0052$&\cellcolor{cyan!25}& \cellcolor{cyan!25} $6.01$   & \cellcolor{cyan!25}$0.450$  \\                            

\hdashline
\multirow{4}{*}{$1024^2$}   & {CIPS-3D \cite{Zhou2021CIPS3DA3}}                     &- & $12.3$        & $0.774$          & - & - & -    &  & -   & -  \\
                            & {StyleNeRF \cite{Gu2022StyleNeRFAS}}                     &$11$ & $8.10 $        & $0.240$          & - & - & -     & & -   & -  \\
                           & {GMPI \cite{zhao2022generative}}                     &$19.4$ & $7.50$     & $0.407$          & $0.75$ & $0.54$ & $0.0070 $    & & n/a   & n/a  \\
                           & \cellcolor{cyan!25}{\textbf{Ours: GMNR}} &\cellcolor{cyan!25}$17.6$ & \cellcolor{cyan!25}$6.58$      & \cellcolor{cyan!25}$0.351 $      & \cellcolor{cyan!25}$0.76$  & \cellcolor{cyan!25}$0.43$  & \cellcolor{cyan!25}$0.0064$ & \cellcolor{cyan!25} & \cellcolor{cyan!25}n/a   & \cellcolor{cyan!25}n/a  \\ 
\bottomrule[0.1em]
\end{tabular}
}
\end{table*}

\noindent\textbf{Ablation Study:} Tab.~\ref{tab:comp_abla} shows the impact of progressively introducing each of our contributions into the baseline on the FFHQ dataset at {$512^2$} resolution. When integrating the proposed $\bm\alpha$-VdR (Sec. \ref{sec:VdR}) into the baseline framework, we observe a consistent improvement in results highlighting the importance of learning view-dependent information during training to render images with diminished artifacts in target poses. Notably, the FID scores reduce from $8.29$ to $7.01$, and the KID scores from $0.454$ to $0.370$. The results are further improved by the introduction of the view-consistency loss $\mathcal{L}_{vc}$ (Sec. \ref{sec:vc_loss}), leading to a consistent gain on all metrics. Our final approach (row 3) that generates 3D-aware view-consistent images achieves an absolute improvement of $1.48$ in terms of FID score over the baseline. 

We further conduct an experiment to ablate the pixel sampling rate in our proposed plane-specific sampling within the $\bm\alpha$-VdR of our GMNR. Here, we ablate the rate from $1\%$ to $6\%$, since $6\%$ is the maximum rate that can be accommodated during our GMNR training under the same batch size setting as the baseline. We observe the results to consistently improve when increasing the sampling rate ($1\%$: $8.32$,  $3\%$: $7.53$, $6\%$: $6.81$ in terms of FID score). As a next step, we also compare our plane-specific sampling with random sampling across planes at the optimal sampling rate ($6\%$).  The random sampling scheme obtains FID and KID scores of $7.68$ and $0.39$. In comparison, our plane-specific sampling-based GMNR improves the results, achieving FID and KID scores of $6.81$ and $0.37$, respectively.

\begin{figure*}[t!]
\centering
\includegraphics[width=1\textwidth]{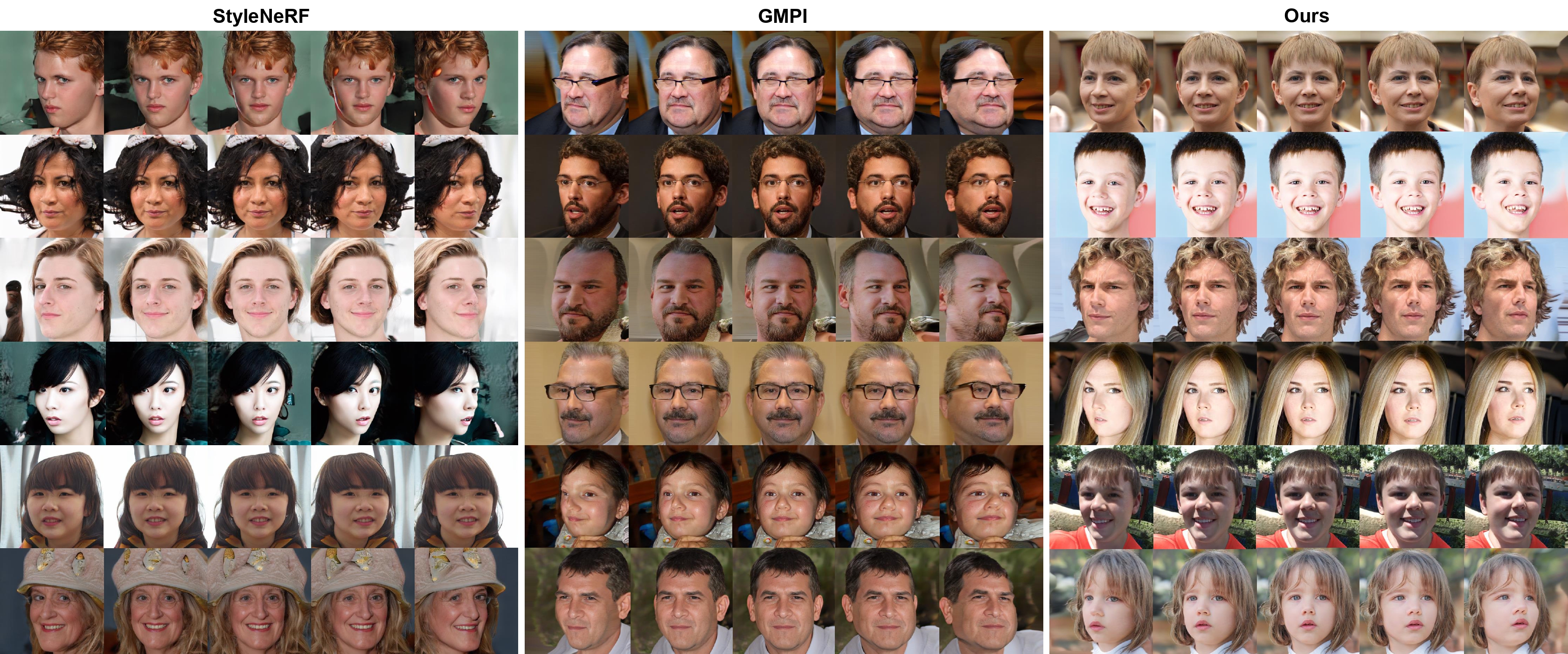}\vspace{0.0cm}
\caption{Qualitative comparison of StyleNeRF \cite{Gu2022StyleNeRFAS}, GMPI \cite{zhao2022generative} and GMNR on FFHQ. For each method, six sets of example images generated at high-resolution ($1024 \times 1024$) are shown. Each set comprises a generated face at canonical view in the center along with four non-frontal views at various angles.  
StyleNeRF-generated images exhibit inconsistent geometry across views, \eg, hair style variation (rows 1, 2, 4 and 5), pupil distortion (row 4), varying artifacts on hat (row 6). On the other hand, GMPI generates near-frontal views of images reasonably well (rows 2 and 4), while artifacts occur at large non-frontal views (rows 1, 3 and 5). Compared to both methods, we observe our GMNR to generate images with enhanced view-consistency and diminished artifacts even at large non-frontal views.\vspace{-0.2cm}
}
\label{fig:qual}
\end{figure*}

\noindent\textbf{Comparison with Existing Approaches:}
Here, we present the comparison of our GMNR with existing works published in literature. Tab.~\ref{tab:compare_sota} presents the comparison on FFHQ and AFHQv2-Cats datasets. 
The results are reported with $256\times 256$, $512 \times 512$ and $1024 \times 1024$ images. 
Both the GMPI and our GMNR approach utilize 96 planes during the test time and do not employ any truncation tricks\cite{brock2018large, karras2019style}. Among existing works, the recent GMPI obtains comparable performance to other methods in literature with faster training, when evaluating on $256^2$ resolution images. For instance, the training time of GMPI for $256^2$ resolution is 3 hours (using 8 Tesla V100 GPUs), excluding the StyleGANv2 pre-training time \cite{zhao2022generative}. Compared to the GMPI training duration (including both StyleGANv2 pre-training and GMPI training), other existing methods including EG3D, GRAM, and StyleNeRF require a longer training time. Moreover, GMPI demonstrates the ability to generate high-resolution images of $1024 \times 1024$ resolutions, where most other existing works struggle to generate. 
When comparing with GMPI approach, our GMNR achieves consistently improved performance on all metrics at different resolutions, without any significant degradation in training time as well as operating at a comparable inference speed. For the high-resolution of $1024 \times 1024$, our GMNR achieves FID score of $6.58$ on the FFHQ dataset, performing favorably against existing published works in the literature while operating at an inference speed of $17.6$ FPS on a single Tesla V100.
Fig.~\ref{fig:qual} presents a comparison of StyleNeRF\cite{Gu2022StyleNeRFAS} and GMPI\cite{zhao2022generative} with GMNR in terms of high-resolution generated image quality at various target views. We show six sets of images that are generated at high-resolution ($1024 \times 1024$) for each method. For each set, we show a generated face at a canonical view in the center as well as the four non-frontal views at different angles. Additional  results are presented in supplementary.

\section{Related Work}
 
Several works have explored rendering 2D multi-view images \cite{garbin2021fastnerf, srinivasan2021nerv, yariv2020multiview, yu2021plenoctrees, zhang2020nerf++} as neural representations of 3D scenes that are differentiable and 3D-aware\cite{atzmon2020sal, chabra2020deep, chen2019learning, eslami2018neural, gropp2020implicit, mescheder2019occupancy, michalkiewicz2019implicit, takikawa2021neural}. 
Based on the scene geometry used, representations can be implicit or explicit.
Explicit representations, such as voxel grid and multiplane images have been employed to render novel views in \cite{sitzmann2019deepvoxels, lombardi2019neural,zhou2018stereo, srinivasan2021nerv, tucker2020single, ghosh2021liveview, li2021mine} for their speed. However, these representations often encounter memory overheads, making it challenging to scale to high-resolutions.
Differently, approaches~\cite{mescheder2019occupancy, Mildenhall2020NeRFRS, Sitzmann2020ImplicitNR, Tancik2020FourierFL} employing implicit 3D representation, such as neural radiance fields~\cite{Mildenhall2020NeRFRS} (NeRF) are memory efficient and can handle complex scenes. However, these approaches struggle with slow rendering limiting the resolution of rendered images. Few works \cite{Devries2021UnconstrainedSG, Liu2020NeuralSV, Martel2021ACORNAC, Peng2020ConvolutionalON, Chan2022EfficientG3} have also explored integrating the merits of these two representations.

In the context of 3D-aware 2D GAN-based image generation, earlier works adopted 3D representations such as, voxel ~\cite{Gadelha20173DSI, Henzler2019EscapingPC, NguyenPhuoc2019HoloGANUL, Wu2016LearningAP, Zhu2018VisualON} and mesh~\cite{Szab2019UnsupervisedG3} for 2D image synthesis. The voxel-based methods are difficult to train on higher-resolution images due to memory requirements of voxel grids and computational overhead of 3D convolution. While \cite{Niemeyer2021GIRAFFERS} partially alleviates this issue via low-resolution rendering followed by 2D upsampling, it struggles to synthesize view-consistent images due to the lack of 3D inductive biases. Furthermore, few works \cite{Chan2021piGANPI, Schwarz2020GRAFGR, Pan2021ASG} utilize neural radiance fields (NeRF) \cite{Mildenhall2020NeRFRS} to generate 3D-aware images, constrained by the slow querying and inefficient GAN training. 
Recent NeRF-based works such as, StyleSDF\cite{OrEl2022StyleSDFH3}, StyleNeRF\cite{Gu2022StyleNeRFAS}, CIPS3D\cite{Zhou2021CIPS3DA3}, GRAM\cite{Deng2022GRAMGR}, VolumeGAN\cite{Xu20223DawareIS} and EG3D\cite{Chan2022EfficientG3} attempt to generate high-resolution images. 
However, these NeRF-based approaches still require significant training time to achieve convergence. In contrast, the recent GMPI method~\cite{zhao2022generative} adopts an MPI-based representation and obtains fast training and rendering speed. However, it struggles to accurately render object shapes at extrapolated views, leading to inconsistent artifacts cross multiple views. Our GMNR addresses these issues by learning a view-dependent representation to generate 3D-aware high-resolution images without degrading the training efficiency and inference speed.
\section{Conclusion}
We introduced an approach, named GMNR, that focuses at efficiently generating 3D-aware high-resolution images that are view-consistent across multiple camera poses. To this end, the proposed GMNR introduces a novel $\bm\alpha$-VdR module that computes the view-dependent representation in an efficient manner by learning viewing direction and position coefficients. Additionally, we employ a view-consistency loss that aims at improving the photometric consistency across multiple views. 
Qualitative and quantitative experiments on three datasets demonstrate the merits of our contributions, leading to favorable performance in terms of  image generation quality and computational efficiency, compared to existing works.


{\small
\bibliographystyle{ieee_fullname}
\bibliography{GMNR}
}

\end{document}